# Credibilistic TOPSIS Model for Evaluation and Selection of Municipal Solid Waste Disposal Methods

Jagannath Roy[1*], Krishnendu Adhikary[1], Samarjit Kar[1]

[1]Department of Mathematics, National Institute of Technology, Durgapur-713209, India

jaga.math23@gmail.com, krish.math23@hotmail.com, dr.samarjitkar@gmail.com

**Abstract:**

Municipal solid waste management (MSWM) is a challenging issue of urban development in developing countries. Each country having different socio-economic-environmental background, might not accept a particular disposal method as the optimal choice. Selection of suitable disposal method in MSWM, under vague and imprecise information can be considered as multi criteria decision making problem (MCDM). In the present paper, TOPSIS (Technique for Order Preference by Similarity to Ideal Solution) methodology is extended based on credibility theory for evaluating the performances of MSW disposal methods under some criteria fixed by experts. The proposed model helps decision makers to choose a preferable alternative for their municipal area. A sensitivity analysis by our proposed model confirms this fact.

## *1. Introduction*

Municipality local body officials or a civic agency plays a vital role in municipal solid waste management (MSWM) since it's become a major issue of urban development. Solid waste disposal and its proper management problem concern about the social, economic, environmental, technical, political factors of a municipality. Due to inadequate consciousness among citizens, local body officials often struggle to make consensus with them in selecting the site for solid waste facility and the method that is to be adopted in the waste disposal Khan et al., 2008. These factors are conflicting in nature. For example, we need to maximize technical reliability, feasibility while net cost per ton, air pollution and emission levels are of minimizing criteria. Many factors (population growth, competitive economy, urbanization and the rise in community living standards) have caused exponential generation rate of municipal solid waste in developing countries Minghua et al., 2009. Mismanagement of municipal solid waste can cause adverse environmental impacts, public health risk and other socio-economic problem Gupta et al., 2015. A long term deficient disposal of municipal solid waste will cause poverty, poor governance and standards of living, major health, environmental and social issues Ogu, 2000. Thus, selection of disposal methods for municipal solid waste is complex decision making problem. This kind of decision making problems are generally ill-structured and often behavioral decision approach. Researchers have shown that human intelligence is typically quite ineffective at solving such problems Promentilla et al., 2006. Decision support tools (DST) such as multi-criteria decision making (MCDM), in this context, have been acknowledged as a key player in MSWM, Charnpratheep et al., 1997; Hung et al., 2007. Moreover, the DST makes an easier communication among decision makers and stakeholders to produce a systematic, transparent, and documented process in decision making.

In this context, MCDM has become a very crucial in management research and decision theory with many methods developed, extended and modified in solving problems in the present and past few decades. Among them TOPSIS model developed by Hwang and Yoon, continues to work satisfactorily across different application areas due to its simple computation and inherent characteristics. Torlak et al., 2010, pointed the basic concept of it is that the chosen alternative should have the shortest distance from the positive-ideal solution (PIS) and longest distance from the negative-ideal solution (NIS). TOPSIS defines an index called closeness coefficient index (cci) to the ideal solution. An alternative with maximum cci value is preferred to be chosen Wang and Chang, 2007.

Earlier researchers dealt with precise and certain information based MCDM methods. Khoo et al., 1999, pointed that a decision support system is based on human knowledge about a specific part of a real or abstract world. The decision can be induced from empirical data from knowledge based opinion. The data and information in decision making are usually imprecise based on experts' subjective judgment where a group of experts is invited to implement the final performance evaluation, ranging from individual judgments to aggregate and rank the alternatives. Many theories, such as, fuzzy sets theory Zadeh, 1965, Dempster–Shafer theory of evidence Shafer, 1976, rough set theory Pawlak, 1982, have been developed to deal with imprecise and incomplete data, human subjective judgment, and real life uncertainty. Suitable extensions of TOPSIS method, based on these theories can be found in the literature.

Few of them are: Jahanshahloo et al., 2006; Chen and Tsao, 2008; Chatterjee and Kar, 2016, and so on. Ye and Li, 2014, introduced the concept of Markowitz's portfolio mean–variance methodology into the traditional TOPSIS methods with fuzzy numbers, and proposed an extended fuzzy TOPSIS method by utilizing the Possibility theory. But a fuzzy event may fail even though its possibility achieves **1**, and may hold even though its necessity is **0**. On the other hand, the fuzzy event must hold if its credibility is **1**, and fail if its credibility is **0**. Thus, there is a disadvantage of using possibility of a fuzzy event Liu, 2004. In order to make more accurate decisions and avoid this drawback, credibility of a fuzzy event is more preferred than possibility of it. So, we utilize credibility of a fuzzy event in TOPSIS model. To the best of our knowledge, there is no work on TOPSIS which uses credibility theory in literature. With these considerations, the paper proposes credibilistic TOPSIS, a novel MCDM model, to assist decision makers for evaluation and selection of the suitable disposal methods for municipal solid waste in urban areas in a country.

The rest of the paper is organized as follows. Section 2 introduces the basic concepts on the fuzzy numbers, and credibility of fuzzy variables. Section 3 presents the proposed credibilistic TOPSIS model based on fuzzy variables. The implementation of the proposed model for evaluating the municipal solid waste disposal methods is provided in section 4. Result discussion is given in section 5. Finally, section 6 concludes the paper.

## 2. Preliminaries

### 2.1. Fuzzy Sets and Fuzzy Numbers

Fuzzy set theory was introduced by Zadeh, (1965), to resolve the impreciseness and ambiguity of human knowledge and judgments. Since its inception fuzzy set theory has been used successfully in data processing by providing mathematical tools to cope with such uncertainties inherited form human thinking and reasoning. Fuzzy MCDM can builds a strong comprehensiveness and reasonableness of the decision-making process Chen et al., 2006. Some applications could be found in Bellman and Zadeh, 1970.

Definition 1. Triangular Fuzzy Numbers (TFN) Chen, 2000,

A Triangular Fuzzy Number, $\xi = (l, m, u)$ is defined by the following membership function

$$\mu_{\tilde{A}}(z) = \begin{cases} 0, & z < l \\ \dfrac{(z-l)}{(m-l)}, & l \leq z \leq m \\ \dfrac{(r-z)}{(r-m)}, & m \leq z \leq u \\ 0, & z > u \end{cases} \quad (1)$$

With $l \leq m \leq u$ is shown in Fig. 1.

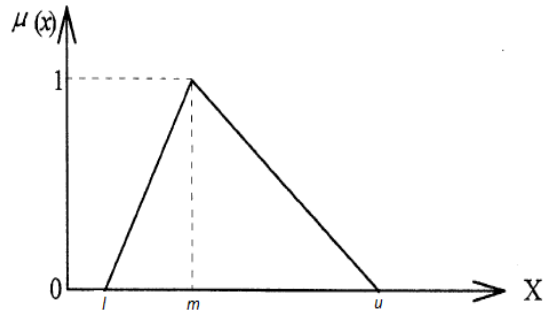

Fig. 1 Triangular Fuzzy Number Chen, 2000

## 2.2. Arithmetic Operation of Fuzzy Numbers

Let $\tilde{A}$ and $\tilde{B}$ be two TFNs parameterized by triplets $(a_1, a_2, a_3)$ and $(b_1, b_2, b_3)$ respectively, then the operational laws of these two TFN are as follows Chen, 2000

1. Addition operation $(+)$ of two TFNs $\tilde{A}$ and $\tilde{B}$:
$$\tilde{A}(+)\tilde{B} = (a_1, a_2, a_3)(+)(b_1, b_2, b_3) = (a_1 + b_1,\ a_2 + b_2,\ a_3 + b_3) \quad (2)$$

2. Subtraction operation $(-)$ of two TFNs $\tilde{A}$ and $\tilde{B}$:
$$\tilde{A}(-)\tilde{B} = (a_1, a_2, a_3)(-)(b_1, b_2, b_3) = (a_1 - b_3,\ a_2 - b_2,\ a_3 - b_1) \quad (3)$$

3. Multiplication operation $(\times)$ of two TFNs $\tilde{A}$ and $\tilde{B}$:
$$\tilde{A}(\times)\tilde{B} = (a_1, a_2, a_3)(\times)(b_1, b_2, b_3) = (a_1 b_1,\ a_2 b_2,\ a_3 b_3) \quad (4)$$

4. Division operation $(\div)$ of two TFNs $\tilde{A}$ and $\tilde{B}$:
$$\tilde{A}(\div)\tilde{B} = (a_1, a_2, a_3)(\div)(b_1, b_2, b_3) = \left(\dfrac{a_1}{b_3}, \dfrac{a_2}{b_2}, \dfrac{a_3}{b_1}\right),\ b_1, b_2, b_3 \neq 0 \quad (5)$$

5. Scalar multiplication in TFNs:
$$k\tilde{A} = (ka_1, ka_2, ka_3), \quad (6)$$

For any real constant $k > 0$.

## 2.3. Aggregation of Fuzzy Numbers

Assume that a decision group has K decision makers, and the fuzzy rating of each decision maker $D_k$ $(k = 1,2,..,K)$ can be represented by a positive TFNs Chen, 2000.

$$R_k = (a_k, b_k, c_k) \ (k = 1,2,..,K)$$

And the aggregated fuzzy rating can be defined as:

$$R = (a, b, c), \qquad k = 1, 2, \ldots, K$$

Where, $a = \min_{1 \leq k \leq K}\{a_k\}$, $b = (b_1 * b_2 * \ldots * b_K)^{\frac{1}{K}}$, and $c = \max_{1 \leq k \leq K}\{c_k\}$.

## 2.5. Expected Value (credibilistic mean) of Fuzzy Variables

**Definition 2 (Liu, 2004).** Let $\xi$ be a fuzzy variable. Then the expected value of $\xi$ is defined by

$$E(\xi) = \int_0^\infty Cr\{\xi \geq r\}dr - \int_{-\infty}^0 Cr\{\xi \leq r\}dr \tag{7}$$

Provided that at least one of the two integrals is finite.

Let $\xi$ be a continuous nonnegative fuzzy variable with membership function $\mu$. If $\mu$ is decreasing on $[0, \infty)$, then $Cr\{\xi \geq x\} = \frac{\mu(x)}{2}$ for any $x > 0$, and $E(\xi) = \frac{1}{2}\int_0^\infty \mu(x)dx$

**Theorem 1.** Let $\xi$ be a continuous fuzzy variable with membership function $\mu$. If its expected value exists, and there is a point $x_0$ such that $x$ is increasing on $(-\infty, x_0)$ and decreasing on $(x_0, \infty)$, then

$$E(\xi) = x_0 + \int_{x_0}^\infty \mu(x)dx - \int_{-\infty}^{x_0} \mu(x)dx \tag{8}$$

Proof: If $x_0 > 0$, then

$$Cr\{\xi \geq r\} = \begin{cases} \frac{1}{2}[1 + 1 - \mu(x)], & if\ 0 \leq r \leq x_0 \\ \frac{1}{2}\mu(x), & if\ r > x_0 \end{cases}$$

And $Cr\{\xi \leq r\} = \frac{1}{2}\mu(x)$

So,

$$E(\xi) = \int_0^{x_0} \left[1 - \frac{1}{2}\mu(x)\right] dx + \int_{x_0}^{\infty} \frac{1}{2}\mu(x)dx - \int_{-\infty}^{x_0} \frac{1}{2}\mu(x)dx$$

$$= x_0 + \int_{x_0}^{\infty} \mu(x)dx - \int_{-\infty}^{x_0} \mu(x)dx$$

If $x_0 < 0$, a similar way may prove the equation. The theorem 1 is proved.

Especially, let $\mu(x)$ be symmetric function on the line $x = x_0$, then $E(\xi) = x_0$.

**Example 1.** Let $\xi$ be a triangular fuzzy variable, i.e., $\xi = (a, b, c)$, then it has an expected value

$$E(\xi) = b + \int_{x_0}^{\infty} \frac{x-c}{c-b} dx - \int_{-\infty}^{x_0} \frac{x-a}{b-a} dx = b + \frac{c-b}{4} + \frac{a-b}{4} = \frac{a+2b+c}{4} \tag{9}$$

## 2.6. Credibilistic Variance of Fuzzy Variables

The variance of a fuzzy variable provides a measure of the spread of the distribution around its expected value. A small value of variance indicates that the fuzzy variable is tightly concentrated around its expected value, and a large value of variance indicates that the fuzzy variable has a wide spread around its expected value.

**Definition 3 (Liu, 2004).** Let $\xi$ be a fuzzy variable with finite expected value $e$. Then the variance of $\xi$ is defined by $V[\xi] = E[(\xi - e)^2]$

Definition 4 (Liu, 2004). Let $\xi$ be a fuzzy variable, and $k$ be a positive number. Then the expected value $E[\xi^k]$ is called the $k$th moment.

**Theorem 2 (Liu, 2004).** Let $\xi$ be a triangular fuzzy variable, i.e., $\xi = (a, b, c)$, then its variance

$$\text{Var}[\xi] = \begin{cases} \frac{33\alpha^3 + 11\alpha\beta^2 + 21\alpha^2\beta - \beta^3}{384\beta}, & \text{if } \alpha > \beta \\ \frac{\alpha^2}{6}, & \text{if } \alpha = \beta \\ \frac{33\beta^3 + 11\alpha^2\beta + 21\alpha\beta^2 - \alpha^3}{384\beta}, & \text{if } \alpha < \beta \end{cases} \tag{10}$$

Where $\alpha = b - a$ and $\beta = c - b$.

Fuzzy Standard deviation, $\sigma = \sqrt{\text{Var}[\xi]}$. \hfill (11)

## 3. The Proposed Fuzzy TOPSIS Based on Credibility Theory

The basic steps of TOPSIS are:

(i) Determine the weights of criteria

(ii) Construction of initial decision matrix

(iii) Computation of normalized decision matrix and weighted normalized decision matrix

(iv) Finding Ideal and Anti-ideal solutions for each criterion

(v) Calculate the distance of each alternative's from the Ideal and Anti-ideal values

(vi) Finally, rank the alternatives based on the closeness coefficient of each alternative's to the ideal solution

Here, we extend TOPSIS model based on credibility theory.

Step 1: *Determine the weighting of evaluation criteria.*

Form a set of Decision Makers (DMs) and identify evaluation criteria. Choose appropriate linguistic variables for the importance weights of criteria and linguistic ratings for alternatives (Table 1). A systematic approach to extend TOPSIS is proposed under fuzzy environment in this section.

Table 1: Fuzzy evaluation scores for the alternatives

| Linguistic Terms | Fuzzy Score | | |
|---|---|---|---|
| Very Poor (VP) | (1, | 1, | 1) |
| Poor (P) | (1, | 1, | 3) |
| Medium Poor (MP) | (1, | 3, | 5) |
| Fair (F) | (3, | 5, | 7) |
| Medium Good (MG) | (5, | 7, | 9) |
| Good (G) | (7, | 9, | 10) |
| Very Good (VG) | (9, | 10, | 10) |

Step 2: *Construct the fuzzy performance/decision matrix*

Choose the appropriate linguistic variables for the alternatives with respect to criteria and form fuzzy decision matrix, $\widetilde{X} = \left(\widetilde{x}_{ij}\right)_{p \times q}$. where $\widetilde{x}_{ij} = (l_{ij}, m_{ij}, u_{ij})$ is the value of the $i^{th}$ alternative according to the $j$th criterion ($i = 1, 2, \ldots, p; j = 1, 2, \ldots, q$).

$$\widetilde{X} = \begin{bmatrix} \widetilde{x}_{11} & \cdots & \widetilde{x}_{1q} \\ \vdots & \ddots & \vdots \\ \widetilde{x}_{p1} & \cdots & \widetilde{x}_{pq} \end{bmatrix}$$

$\widetilde{x}_{ij}^k = (l_{ij}^k, m_{ij}^k, u_{ij}^k)$ is the rating of the alternative $A_i$ w.r.t. criterion $C_j$ evaluated by expert $k$ and

$$l_{ij} = \min_{1 \le k \le K}\{l_{ij}^k\}, \quad m_{ij} = \left(m_{ij}^k * m_{ij}^k * \ldots * m_{ij}^k\right)^{\frac{1}{K}}, \quad \text{and} \quad u_{ij} = \max_{1 \le k \le K}\{u_{ij}^k\}.$$

where $p$ denotes the number of the alternatives, $q$ stands for the total number of criteria.

Step 3. *Normalize the elements from the initial fuzzy decision matrix ($\widetilde{X}$).*

$$\widetilde{N} = \begin{bmatrix} \widetilde{n}_{11} & \cdots & \widetilde{n}_{1q} \\ \vdots & \ddots & \vdots \\ \widetilde{n}_{p1} & \cdots & \widetilde{n}_{pq} \end{bmatrix}$$

The elements of the fuzzy normalized matrix ($\widetilde{N}$) are computed using the equations (12) and (13)

- For Benefit type criteria (maximizing type)
  - $\widetilde{n}_{ij} = (l'_{ij}, m'_{ij}, u'_{ij}) = \left(\dfrac{l_{ij}}{u_j^*}, \dfrac{m_{ij}}{u_j^*}, \dfrac{u_{ij}}{u_j^*}\right)$ (12)

- b) For Cost type criteria (minimizing type)
  - $\widetilde{n}_{ij} = (l'_{ij}, m'_{ij}, u'_{ij}) = \left(\dfrac{l_j^-}{u_{ij}}, \dfrac{l_j^-}{m_{ij}}, \dfrac{l_j^-}{l_{ij}}\right)$ (13)

Let $x_{ij}, x_j^*$ and $x_j^-$ are the elements from the initial fuzzy decision matrix ($\widetilde{X}$), for which $x_j^*$ and $x_j^-$ are defined

$u_j^* = \max_{1 \le i \le m}(u_{ij}), j = 1, 2, \ldots, n$

And

$l_j^- = \min_{1 \le i \le m}(l_{ij}), j = 1, 2, \ldots, n$

The normalization approach mentioned above is to make the ranges of normalized triangular fuzzy numbers belong to [0, 1].

Step 4. *Construct the credibilistic mean value matrix*

According to equation (9), we can get the expected value (credibilistic mean) of triangular fuzzy number

$$E[\widetilde{N}] = \begin{bmatrix} E(\widetilde{n}_{11}) & \cdots & E(\widetilde{n}_{1q}) \\ \vdots & \ddots & \vdots \\ E(\widetilde{n}_{p1}) & \cdots & E(\widetilde{n}_{pq)} \end{bmatrix}$$

Step 5. *Construct the credibilistic standard deviation matrix*

Calculate the credibilistic standard deviation matrix $\sigma(\widetilde{N})$ using equations (10)-(11) about the fuzzy normalized matrix $\widetilde{N}$ which is given by

$$\sigma(\widetilde{N}) = \begin{bmatrix} \sigma(\widetilde{n}_{11}) & \cdots & \sigma(\widetilde{n}_{1q}) \\ \vdots & \ddots & \vdots \\ \sigma(\widetilde{n}_{p1}) & \cdots & \sigma(\widetilde{n}_{pq)} \end{bmatrix}$$

Step 6. *Calculating the Positive Ideal Solution (PIS) and Negative Ideal Solution (NIS) about credibilistic mean value matrix*

Next, we determine the PIS $E(\widetilde{N})^+$ and NIS $E(\widetilde{N})^-$ about the *credibilistic mean value matrix* $E[\widetilde{N}]$ for the decision maker as:

$$E(\widetilde{N})^+ = \left(E(\widetilde{n}_1)^+, E(\widetilde{n}_2)^+, \ldots, E(\widetilde{n}_q)^+\right) \quad (14.a)$$

$$E(\widetilde{N})^- = \left(E(\widetilde{n}_1)^-, E(\widetilde{n}_2)^-, \ldots, E(\widetilde{n}_q)^-\right) \quad (14.b)$$

$$E(\widetilde{n}_j)^+ = \max_{1 \leq i \leq p}\{E(\widetilde{n}_{ij})\} \quad (14.c)$$

and

$$E(\widetilde{n}_j)^- = \min_{1 \leq i \leq p}\{E(\widetilde{n}_{ij})\} \quad j = 1, 2, \ldots, q \quad (14.d)$$

Step 7. *Determine the PIS and NIS about credibilistic standard deviation matrix*

Furthermore, we identify PIS $\sigma(\widetilde{N})^+$ and NIS $\sigma(\widetilde{N})^-$ about the expected value matrix $E[\widetilde{N}]$ for the decision maker as:

$$\sigma(\widetilde{N})^+ = \left(\sigma(\widetilde{n}_1)^+, \sigma(\widetilde{n}_2)^+, \ldots, \sigma(\widetilde{n}_q)^+\right) \quad (15.a)$$

$$\sigma(\widetilde{N})^- = \left(\sigma(\widetilde{n}_1)^-, \sigma(\widetilde{n}_2)^-, \ldots, \sigma(\widetilde{n}_q)^-\right) \tag{15.b}$$

$$\sigma(\widetilde{n}_j)^+ = \min_{1 \leq i \leq p}\{\sigma(\widetilde{n}_{ij})\} \tag{15.c}$$

and

$$\sigma(\widetilde{n}_j)^- = \max_{1 \leq i \leq p}\{\sigma(\widetilde{n}_{ij})\} \qquad j = 1, 2, \ldots, q \tag{15.d}$$

Step 8. *Finding the separation measure of each alternative's credibilistic mean value from PIS and NIS*

By using the $q$-dimensional Euclidean distance, the separation measures of each alternative's expected value from the PIS $E(\widetilde{N})^+$ and expected value from the PIS $\sigma(\widetilde{N})^+$ are given as:

$$d_i\left(E(\widetilde{N})^+\right) = \left\{\sum_{j=1}^q \left[\left(E(\widetilde{n}_j)^+ - E(\widetilde{n}_{ij})\right) w_j\right]^2\right\}^{\frac{1}{2}} \tag{16}$$

$$d_i(E(\widetilde{N})^-) = \left\{\sum_{j=1}^q \left[\left(E(\widetilde{n}_j)^- - E(\widetilde{n}_{ij})\right) w_j\right]^2\right\}^{\frac{1}{2}} \tag{17}$$

Step 9. *Compute the distance measure of each alternative's credibilistic standard deviation from PIS and NIS*

By using the $q$-dimensional Euclidean distance, the distance measures of each alternative's expected value from the NIS $E(\widetilde{N})^-$ and expected value from the NIS $\sigma(\widetilde{N})^-$ are given as:

$$d_i\left(\sigma(\widetilde{N})^+\right) = \left\{\sum_{j=1}^q \left[\left(\sigma(\widetilde{n}_j)^+ - \sigma(\widetilde{n}_{ij})\right) w_j\right]^2\right\}^{\frac{1}{2}} \tag{18}$$

$$d_i(\sigma(\widetilde{N})^-) = \left\{\sum_{j=1}^q \left[\left(\sigma(\widetilde{n}_j)^- - \sigma(\widetilde{n}_{ij})\right) w_j\right]^2\right\}^{\frac{1}{2}} \tag{19}$$

Step 10. *Compute the closeness coefficient of alternative about its credibilistic mean value and credibilistic standard deviation*

A closeness coefficient is defined to determine the ranking order of all alternatives once $d_i\left(E(\widetilde{N})^+\right), d_i\left(\sigma(\widetilde{N})^+\right), d_i(E(\widetilde{N})^-)$, and $d_i(\sigma(\widetilde{N})^-)$ of each alternative $A_i$ are found. Now, an alternative $A_i$, is good if it is closest to the ideal solution and farthest from the anti-ideal one. Thus for making optimal choice we need to find relative closeness coefficient of each alternatives. Here, in this case, relative closeness coefficient about its expected value and standard deviation are defined as

$$CC_i\left(E(\widetilde{N})\right) = \frac{d_i(E(\widetilde{N})^-)}{d_i\left(E(\widetilde{N})^+\right)+d_i(E(\widetilde{N})^-)} \tag{20}$$

and

$$CC_i\left(\sigma(\widetilde{N})\right) = \frac{d_i(\sigma(\widetilde{N})^-)}{d_i\left(\sigma(\widetilde{N})^+\right)+d_i(\sigma(\widetilde{N})^-)} \tag{21}$$

Step 11. *Ranking of the alternatives according to the final relative closeness coefficient*

Then the final relative closeness coefficient of alternative $A_i$ is given as

$$CC(A_i) = \sqrt{CC_i\left(E(\widetilde{N})\right) \times CC_i\left(\sigma(\widetilde{N})\right)} \tag{22}$$

The alternatives are ranked in decreasing value of integrated relative closeness coefficients.

## 4. *Implementation of the Proposed Method for Ranking Solid Waste Disposal Methods*

In this section, a numerical example adopted from Ekmekcioglu et al., 2010 is provided to illustrate the application of the proposed method for the site selection in solid waste disposal methods. Criteria and alternative disposal methods are briefly described in Table 2 and Table 3 respectively.

Table 2: Criteria and their brief description

| Criteria | Criteria Name | Type |
|---|---|---|
| $C1$ | Technical reliability | Maximizing |
| $C2$ | Feasibility | Maximizing |
| $C3$ | Separation of waste materials | Maximizing |
| $C4$ | Waste recovery | Maximizing |
| $C5$ | Energy recovery | Maximizing |
| $C6$ | Net cost per ton | Minimizing |
| $C7$ | Air Pollution Control | Minimizing |
| $C8$ | Emission levels | Minimizing |
| $C9$ | Surface water dispersed releases | Minimizing |
| $C10$ | Number of employees | Minimizing |

Table 3: Alternatives and their brief description

| Alternative | Disposal methods | Brief Description |
|---|---|---|
| $A1$ | Landfilling | Landfilling is the most commonly employed method in MSWM. It integrates an engineering based method for disposal of solid waste on land in such a manner that environmental hazards are reduced while spreading the solid waste in thin layers. |
| $A2$ | Composting | The four tasks are crucial to for designing a modern MSW composting system: collection, contaminant separation, sizing and mixing, and biological decomposition. |
| $A3$ | RDF combustion | Refuse-derived fuel (RDF) combustion is developed to avoid the immediate burn of MSW. Instead of immediate burning, in this system, MSW is turned into a transportable, storable fuel. While conventional incineration demands little sorting or processing of the waste, in RDF production the waste may undergo a number of preprocessing stages. |
| $A4$ | Conventional incineration | Conventional incineration which is extremely effective a waste management and volume reduction technique since the 1890s. In terms of waste processing, conventional incineration is a relatively simple option, with unsorted waste being fed into a furnace and, by burning, reduced to one-tenth of its original volume. |

Table 4: Weight priorities of criteria set by experts

| Criteria: | $C1$ | $C2$ | $C3$ | $C4$ | $C5$ | $C6$ | $C7$ | $C8$ | $C9$ | $C10$ |
|---|---|---|---|---|---|---|---|---|---|---|
| Criteria Weights: | 0.020 | 0.014 | 0.136 | 0.200 | 0.240 | 0.033 | 0.146 | 0.121 | 0.003 | 0.087 |

Table 5: Decision makers' Opinion for evaluation of alternatives

|    |     | C1 | C2 | C3 | C4 | C5 | C6 | C7 | C8 | C9 | C10 |
|----|-----|----|----|----|----|----|----|----|----|----|-----|
| A1 | DM1 | P  | MG | F  | MP | MG | VG | F  | G  | P  | MP  |
|    | DM2 | MP | G  | F  | P  | F  | VG | F  | MG | VP | P   |
|    | DM3 | MP | MG | F  | F  | F  | VG | MG | F  | P  | MP  |
| A2 | DM1 | F  | VP | F  | F  | F  | VG | MG | G  | MP | P   |
|    | DM2 | MP | MP | F  | P  | MG | VG | F  | MG | VP | MP  |
|    | DM3 | F  | VP | F  | F  | G  | VG | F  | F  | MP | MP  |
| A3 | DM1 | G  | VP | G  | G  | G  | VP | F  | F  | MG | F   |
|    | DM2 | MG | P  | MG | MG | G  | VP | MG | MG | G  | MG  |
|    | DM3 | G  | VP | G  | G  | G  | VP | F  | MG | G  | F   |
| A4 | DM1 | F  | P  | F  | P  | P  | VP | P  | P  | F  | F   |
|    | DM2 | F  | P  | F  | P  | P  | G  | P  | MP | F  | F   |
|    | DM3 | F  | P  | F  | P  | P  | G  | P  | MP | F  | F   |

*Step 1*: Weight priorities of the criteria given by the decision maker's is shown in table 4.

*Step 2:* Decision maker's opinion for evaluation of alternatives are presented in table 5. Initialize the decision matrix we have used in section 2.3 is shown in table 6.

*Step 3*. For normalization we have used equation (12) and (13) for maximization or minimization criteria respectively shown in table 7.

*Step 4*. The credibilistic mean value matrix about the normalized fuzzy decision matrix is calculated using equation (9), shown in table 8.

Table 6: Aggregated Initial Fuzzy Decision matrix ($\widetilde{X}$)

|     | A1 | A2 | A3 | A4 |
|-----|----|----|----|----|
| C1  | (1.000, 2.080, 5.000) | (1.000, 4.217, 7.000) | (5.000, 8.277, 10.00) | (3.000, 5.000, 7.000) |
| C2  | (5.000, 7.612, 10.00) | (1.000, 1.442, 5.000) | (1.000, 1.000, 3.000) | (1.000, 1.000, 3.000) |
| C3  | (3.000, 5.000, 7.000) | (3.000, 5.000, 7.000) | (5.000, 8.277, 10.00) | (3.000, 5.000, 7.000) |
| C4  | (1.000, 2.466, 7.000) | (1.000, 2.924, 7.000) | (5.000, 8.277, 10.00) | (1.000, 1.000, 3.000) |
| C5  | (3.000, 5.593, 9.000) | (3.000, 6.804, 10.00) | (7.000, 9.000, 10.00) | (1.000, 1.000, 3.000) |
| C6  | (9.000, 10.00, 10.00) | (9.000, 10.00, 10.00) | (1.000, 1.000, 1.000) | (1.000, 4.327, 10.00) |
| C7  | (3.000, 5.593, 9.000) | (3.000, 5.593, 9.000) | (3.000, 5.593, 9.000) | (1.000, 1.000, 3.000) |
| C8  | (3.000, 6.804, 10.00) | (3.000, 6.804, 10.00) | (3.000, 6.257, 9.000) | (1.000, 2.080, 5.000) |
| C9  | (1.000, 1.000, 3.000) | (1.000, 2.080, 5.000) | (5.000, 8.277, 10.00) | (3.000, 5.000, 7.000) |
| C10 | (1.000, 2.080, 5.000) | (1.000, 2.080, 5.000) | (3.000, 5.593, 9.000) | (3.000, 5.000, 7.000) |

*Step 5*. The credibilistic standard deviation matrix about the normalized fuzzy decision matrix is formed using equation (10) and (11).

Table 7: Normalized Fuzzy decision Matrix ($\widetilde{N}$)

|  | A1 | A2 | A3 | A4 |
|---|---|---|---|---|
| $C1$ | (0.100, 0.208, 0.500) | (0.100, 0.422, 0.700) | (0.500, 0.828, 1.000) | (0.300, 0.500, 0.700) |
| $C2$ | (0.500, 0.761, 1.000) | (0.100, 0.144, 0.500) | (0.100, 0.100, 0.300) | (0.100, 0.100, 0.300) |
| $C3$ | (0.300, 0.500, 0.700) | (0.300, 0.500, 0.700) | (0.500, 0.828, 1.000) | (0.300, 0.500, 0.700) |
| $C4$ | (0.100, 0.247, 0.700) | (0.100, 0.292, 0.700) | (0.500, 0.828, 1.000) | (0.100, 0.100, 0.300) |
| $C5$ | (0.300, 0.559, 0.900) | (0.300, 0.680, 1.000) | (0.700, 0.900, 1.000) | (0.100, 0.100, 0.300) |
| $C6$ | (0.100, 0.100, 0.111) | (0.100, 0.100, 0.111) | (1.000, 1.000, 1.000) | (0.100, 0.231, 1.000) |
| $C7$ | (0.111, 0.179, 0.333) | (0.111, 0.179, 0.333) | (0.111, 0.179, 0.333) | (0.333, 1.000, 1.000) |
| $C8$ | (0.100, 0.147, 0.333) | (0.100, 0.147, 0.333) | (0.111, 0.160, 0.333) | (0.200, 0.481, 1.000) |
| $C9$ | (0.333, 1.000, 1.000) | (0.200, 0.481, 1.000) | (0.100, 0.121, 0.200) | (0.143, 0.200, 0.333) |
| $C10$ | (0.200, 0.481, 1.000) | (0.200, 0.481, 1.000) | (0.111, 0.179, 0.333) | (0.143, 0.200, 0.333) |

Table 8: Credibilistic Mean value Matrix $E(\widetilde{N})$

|  | $C1$ | $C2$ | $C3$ | $C4$ | $C5$ | $C6$ | $C7$ | $C8$ | $C9$ | $C10$ |
|---|---|---|---|---|---|---|---|---|---|---|
| $A1$ | 0.254 | 0.756 | 0.500 | 0.323 | 0.580 | 0.103 | 0.201 | 0.182 | 0.833 | 0.540 |
| $A2$ | 0.411 | 0.222 | 0.500 | 0.346 | 0.665 | 0.103 | 0.201 | 0.182 | 0.540 | 0.540 |
| $A3$ | 0.789 | 0.150 | 0.789 | 0.789 | 0.875 | 1.000 | 0.201 | 0.191 | 0.135 | 0.201 |
| $A4$ | 0.500 | 0.150 | 0.500 | 0.150 | 0.150 | 0.391 | 0.833 | 0.540 | 0.219 | 0.219 |

Table 9: Credibilistic Standard Deviation Matrix $\sigma(\widetilde{N})$

|  | $C1$ | $C2$ | $C3$ | $C4$ | $C5$ | $C6$ | $C7$ | $C8$ | $C9$ | $C10$ |
|---|---|---|---|---|---|---|---|---|---|---|
| $A1$ | 0.097 | 0.102 | 0.082 | 0.148 | 0.129 | 0.003 | 0.052 | 0.059 | 0.195 | 0.183 |
| $A2$ | 0.123 | 0.109 | 0.082 | 0.140 | 0.144 | 0.003 | 0.052 | 0.059 | 0.183 | 0.183 |
| $A3$ | 0.108 | 0.059 | 0.108 | 0.108 | 0.065 | 0.000 | 0.052 | 0.056 | 0.025 | 0.052 |
| $A4$ | 0.082 | 0.059 | 0.082 | 0.059 | 0.059 | 0.238 | 0.195 | 0.183 | 0.045 | 0.045 |

Table 10: PIS and NIS about the Credibilistic Mean value Matrix

|  | $C1$ | $C2$ | $C3$ | $C4$ | $C5$ | $C6$ | $C7$ | $C8$ | $C9$ | $C10$ |
|---|---|---|---|---|---|---|---|---|---|---|
| $E(N)^+$ | 0.789 | 0.756 | 0.789 | 0.789 | 0.875 | 1.000 | 0.833 | 0.540 | 0.833 | 0.540 |
| $E(N)^-$ | 0.254 | 0.150 | 0.500 | 0.150 | 0.150 | 0.103 | 0.201 | 0.182 | 0.135 | 0.201 |

Table 11: PIS and NIS about the Standard deviation Matrix

|  | C1 | C2 | C3 | C4 | C5 | C6 | C7 | C8 | C9 | C10 |
|---|---|---|---|---|---|---|---|---|---|---|
| $\sigma(N)^+$ | 0.082 | 0.059 | 0.082 | 0.059 | 0.059 | 0.000 | 0.052 | 0.056 | 0.025 | 0.045 |
| $\sigma(N)^-$ | 0.123 | 0.109 | 0.108 | 0.148 | 0.144 | 0.238 | 0.195 | 0.183 | 0.195 | 0.183 |

Table 12: Separation Measure of each alternative's Credibilistic mean value from PIS and NIS

|  | A1 | A2 | A3 | A4 |
|---|---|---|---|---|
| $d_i\left(E(\tilde{N})^+\right)$ | 0.163 | 0.153 | 0.106 | 0.223 |
| $d_i\left(E(\tilde{N})^-\right)$ | 0.113 | 0.133 | 0.222 | 0.103 |

Table 13: Separation Measure of each alternative's standard deviation matrix from PIS and NIS

|  | A1 | A2 | A3 | A4 |
|---|---|---|---|---|
| $d_i\left(\sigma(\tilde{N})^+\right)$ | 0.027 | 0.029 | 0.011 | 0.027 |
| $d_i\left(\sigma(\tilde{N})^-\right)$ | 0.027 | 0.027 | 0.036 | 0.030 |

*Step 6:* Then calculate the Positive Ideal Solution (PIS) and Negative Ideal Solution (NIS) about credibilistic mean value matrix by using the equations (14.a), (14.b), (14.c), (14.d); given in table 10.

*Step 7.* Next, PIS and NIS about credibilistic standard deviation matrix is calculated according to equations (15.a), (15.b), (15.c), (15.d); shown in table 11.

*Step 8.* Find the distance measure of each alternative's credibilistic mean value from PIS and NIS by using equations (16), (17) presented in table 12.

*Step 9:* Table 13 is formed after computing the distance measure of each alternative's credibilistic standard deviation from PIS and NIS using equations (18) and (19).

*Step 10*: Closeness coefficient of alternative about its expected value and standard deviation are find by equations (20) and (21).

Table 14: The closeness coefficient and ranking of each alternative

| Alternatives | $CC_i\left(E(\widetilde{N})\right)$ | $CC_i\left(\sigma(\widetilde{N})\right)$ | $CC(A_i)$ | Rank |
|---|---|---|---|---|
| $A1$ | 0.409 | 0.500 | 0.452 | 3 |
| $A2$ | 0.465 | 0.485 | 0.475 | 2 |
| $A3$ | 0.676 | 0.771 | 0.722 | 1 |
| $A4$ | 0.316 | 0.525 | 0.407 | 4 |

*Step 11*: The alternatives are ranked in decreasing value of final relative closeness coefficients by using equation (22) shown in the table 14.

## 5. Result Discussion

### 5.1 Comparison with other Models

In order to verify the validity of our proposed method, we perform a comparison of our proposed method with two other previous methods including fuzzy TOPSIS Chen et al., 2006 and Possibilistic TOPSIS Ye and Li, 2014 which also deal with fuzzy numbers. The results are shown as follows:

Table 15: Comparison with other models

|  | Fuzzy TOPSIS (Chen, 2006) | | Possibilistic TOPSIS (Ye and Li, 2014) | | Proposed TOPSIS model | |
|---|---|---|---|---|---|---|
|  | $CC_i$ | Rank | $CC_i$ | Rank | $CC_i$ | Rank |
| $A1$ | 0.5570 | 3 | 0.2180 | 3 | 0.452 | 3 |
| $A2$ | 0.5701 | 2 | 0.1871 | 4 | 0.475 | 2 |
| $A3$ | 0.7429 | 1 | 0.2746 | 1 | 0.722 | 1 |
| $A4$ | 0.3178 | 4 | 0.2384 | 2 | 0.407 | 4 |

It is clear from Table 15 that the three methods have the similar results. Note that **$A3$** is best alternative according to all three models under fixed preference of criteria weights. These shows the method we proposed in this paper is reasonable.

### 5.2 Sensitivity Analysis

An MCDM method is dependent on the weight priorities of the criteria that is, on the relative importance involved to particular criteria. Occasionally, the final evaluation and ranking change whenever there is a very small variation in the relative preference of the criteria. As a result of which, a good MCDM method should act as a sensitive to the change in criteria weights.

In this research work, a sensitivity analysis was performed to assess how changes in the weights assigned to the criteria would change the ranking of the alternatives. Different weight priorities are given to the criteria (see Table 16) corresponding to different *scenarios (1)-(6)*.

Table 16: weight priorities of criteria for different scenario

|       | Scenario 1 | Scenario 2 | Scenario 3 | Scenario 4 | Scenario 5 | Scenario 6 |
|-------|------------|------------|------------|------------|------------|------------|
| $C1$  | 0.1117     | 0.0613     | 0.1458     | 0.1692     | 0.1024     | 0.2367     |
| $C2$  | 0.0061     | 0.1510     | 0.0495     | 0.0512     | 0.2161     | 0.1726     |
| $C3$  | 0.1446     | 0.1456     | 0.0982     | 0.1639     | 0.0432     | 0.0974     |
| $C4$  | 0.1590     | 0.0361     | 0.1356     | 0.0490     | 0.0626     | 0.1424     |
| $C5$  | 0.1156     | 0.0264     | 0.1728     | 0.1870     | 0.0345     | 0.1115     |
| $C6$  | 0.1290     | 0.1107     | 0.1861     | 0.0704     | 0.0323     | 0.0211     |
| $C7$  | 0.1265     | 0.2133     | 0.1062     | 0.0396     | 0.2063     | 0.0666     |
| $C8$  | 0.0668     | 0.0756     | 0.0269     | 0.0505     | 0.1376     | 0.0342     |
| $C9$  | 0.1116     | 0.1301     | 0.0290     | 0.1240     | 0.1305     | 0.0510     |
| $C10$ | 0.0291     | 0.0497     | 0.0500     | 0.0952     | 0.0344     | 0.0666     |

The resulting rankings are given in Table 17. The results show that assigning different weights (priorities) to the criteria leads to different rankings, i.e., the proposed model is sensitive to these weights, which is necessary for any MCDM models. On comparison of all scenarios we find *A3* is best choice except in *scenario 5*. However, *scenario 1* and *scenario 4* provide same ranking order though weight priorities have large change. From scenario (2) and (3), it is clear that a slight change in criteria weights will lead to alter *A2* and *A3* as second best choice. Similar arguments could be done for other pairwise comparisons among the scenarios. These comparisons confirms that using different weights to the criteria under consideration may help to choose the best design alternative in different context if needed.

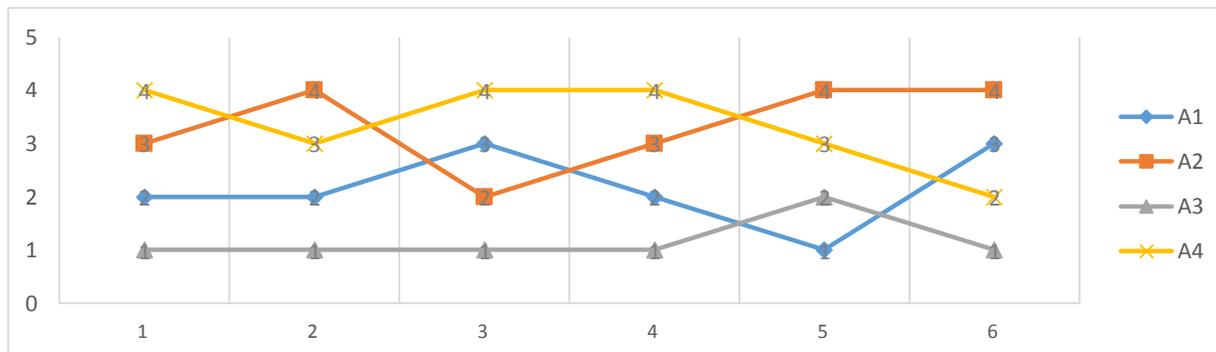

Fig.2. Ranking of the alternatives in various scenario

Table 17: The closeness coefficient and ranking of each alternative in different scenario

|    | Scenario 1 Ranking | Scenario 2 Ranking | Scenario 3 Ranking | Scenario 4 Ranking | Scenario 5 Ranking | Scenario 6 Ranking |
|----|---|---|---|---|---|---|
| A1 | 2 | 2 | 3 | 2 | 1 | 3 |
| A2 | 3 | 4 | 2 | 3 | 4 | 4 |
| A3 | 1 | 1 | 1 | 1 | 2 | 1 |
| A4 | 4 | 3 | 4 | 4 | 3 | 2 |

## *6. Conclusion*

This study proposes the credibilistic TOPSIS model based on expected value (Credibilistic mean value) operator for a fuzzy variable to facilitate a more precise analysis of the alternatives, considering several criteria in imprecise environment. The proposed method is applied successfully to select the most preferable disposal method(s) for municipal solid waste for imprecise data. Also, this model provides the expected optimal choice for disposal method(s) after effectively avoiding vague and ambiguous judgments.

In many practical MCDM problems different relative weights of criteria must be taken into consideration since these are influenced by the socio-economic-environmental-technical condition of a country (for municipal solid waste disposal). So, a particular disposal method might not be accepted as optimal solution to all countries. It is evident that the sensitivity analysis performed (Table 16 & 17) through the proposed model shows that *A3* is not the optimal choice in all cases (scenario 4 & 5). In future, credibilistic TOPSIS would produce interesting hybrid MCDM methods with the combination of other MCDM techniques like, ANP, DEMATEL and Shannon Entropy etc.

**Acknowledgement:** The first author (Jagannath Roy) would like to thank Department of Science and Technology, India, for their help and supports in this research work under INSPIRE program.

**Conflict of interest:** The authors declare that there is no conflict of interest regarding the publication of this paper.

12. Zadeh, L. (1965). Fuzzy sets. *Information and Control* 8 (3): 338–53.

13. Shafer, G. (1976). A mathematical theory of evidence (Vol. 1, pp. xiii+-297). *Princeton: Princeton university press*.

14. Pawlak Z. (1982). Rough sets. *International Journal of Computing and Information Sciences*, 11(5), 341–356.

15. Torlak, G., Sevkli, M., Sanal, M., & Zaim, S. (2011). Analyzing business competition by using fuzzy TOPSIS method: An example of Turkish domestic airline industry. *Expert Systems with Applications*, 38(4), 3396-3406.

16. Wang, T. C., & Chang, T. H. (2007). Application of TOPSIS in evaluating initial training aircraft under a fuzzy environment. *Expert Systems with Applications*, 33(4), 870-880.

17. Jahanshahloo, G. R., Lotfi, F. H., & Izadikhah, M. (2006). Extension of the TOPSIS method for decision-making problems with fuzzy data. *Applied Mathematics and Computation*, 181(2), 1544-1551.

18. Chen, T. Y., & Tsao, C. Y. (2008). The interval-valued fuzzy TOPSIS method and experimental analysis. *Fuzzy Sets and Systems*, 159(11), 1410-1428.

19. Chatterjee, K., & Kar, S. (2016). Multi-criteria analysis of supply chain risk management using interval valued fuzzy TOPSIS. *OPSEARCH*, 1-26.

20. Ye, F., & Li, Y. (2014). An extended TOPSIS model based on the possibility theory under fuzzy environment. *Knowledge-Based Systems*, 67, 263-269.

21. Liu, B. D. (2004). Uncertain theory: An introduction to its axiomatic foundation. Berlin: Springer-Verlag.

22. Chen, C. T. (2000). Extensions of the TOPSIS for group decision-making under fuzzy environment. *Fuzzy sets and systems*, 114(1), 1-9.

23. Chen, C. T., Lin, C. T., & Huang, S. F. (2006). A fuzzy approach for supplier evaluation and selection in supply chain management. *International journal of production economics*, 102(2), 289-301.

24. Bellman, R. E., & Zadeh, L. A. (1970). Decision-making in a fuzzy environment. *Management science*, 17(4), B-141.

25. Ekmekçioğlu, M., Kaya, T., & Kahraman, C. (2010). Fuzzy multicriteria disposal method and site selection for municipal solid waste. *Waste Management*, 30(8), 1729-1736.
20